# Middle Architecture Criteria

John Beverley[1,2,3,*], Giacomo De Colle[1,2], Mark Jensen[3], Carter Benson[1,2], and Barry Smith[1,2,3]

[1] *Department of Philosophy, University at Buffalo, NY, USA*
[2] *National Center for Ontological Research, NY, USA*
[3] *Institute for Artificial Intelligence and Data Science, University at Buffalo, NY, USA*
[4] *U.S. Customs and Border Protection, USA*

**Abstract**
Mid-level ontologies are used to integrate data across disparate domains using vocabularies more specific than top-level ontologies and more general than domain-level ontologies. There are no clear, defensible criteria for determining whether a given ontology should count as mid-level, because we lack a rigorous characterization of what the middle level of generality is supposed to contain. Attempts to provide such a characterization have failed, we believe, because they have focused on the goal of specifying what is characteristic of those single ontologies that have been advanced as mid-level ontologies. Unfortunately, single ontologies of this sort are generally a mixture of top- and mid-level, and sometimes even of domain-level terms. To gain clarity, we aim to specify conditions for membership in what we call the middle architecture, which consists solely of mid-level ontologies.

## 1. Introduction

Ontologists distinguish top-, mid-, and bottom-level ontologies [1]. Top-level ontologies (also known as "upper" or "foundational" ontologies) are implemented using languages composed of the most general terms and relational expressions, reflecting broad areas such as mereology, space, time, and so forth [2]. Bottom-level ontologies (also known as "domain" ontologies) are implemented in domain-specific languages, where a domain is understood to be a collection of entities of interest to a certain community or discipline [3], such as occupations, proteins, cats, clouds, legal entities, and so on. Mid-level ontologies are implemented in languages composed of terms and relational expressions that are more specific than what would be found in the top level, yet more general than what would be found at the bottom [4, 5, 6].[2]

While intuitive, the preceding provides limited guidance regarding what counts as a top-, domain-, or mid-level ontology; providing such guidance is no mere intellectual exercise. Growing interest in enterprise ontology solutions has led to a need for standardized, domain- and mid-level ontologies extending from vetted, established, top-level ontologies [8]. Simple analogies illustrate why. Where top-level ontologies are analogous to programming languages such as Python; mid-level ontologies are analogous to programming language libraries such as Pandas or NumPy [1]. Just as developers leverage libraries to avoid having to start from scratch when writing software applications, so ontologists operating at the domain level benefit by leveraging mid-level ontologies. Motivation of this sort has, as a recent example, led to an on-going effort sponsored by the Institute of Electrical and Electronics Engineers Standards Association (IEEE) aimed at identifying requirements for mid-level ontologies [9].

---

[*] Corresponding author.
✉ johnbeve@buffalo.edu (J. Beverley); gdecolle@buffalo.edu (G. De Colle); mark.p.jensen@cbp.dhs.gov (M. Jensen); carterbe@buffalo.edu (C. Benson); ifomis@gmail.com (B. Smith)

[2] This is, of course, not the only classification of ontologies; see [7] for discussion of top-level, domain, task, and application ontologies; see [3] for discussion of top-level, reference, and a certain type of application ontology.

While progress has been made on identifying criteria for what counts as a top-level ontology [2], mid-level ontology criteria have proven particularly elusive. Given known challenges to constructing such criteria [10, 11] – some of which are discussed below – we focus here on conditions for membership in *ontology architectures*, which for our purposes are classifications of ontologies based on levels of generality broadly understood (**Figure 1**).[3] By providing individually necessary and jointly sufficient conditions for membership in an architecture, we can maintain that, for example, members of the middle architecture are mid-level ontologies, without being committed to all mid-level ontologies being members of the middle architecture. In other words, rather than attempt to identify features common to all mid-level ontologies, we identify mid-level ontologies in terms of important features they exhibit.

Ontology architectures track the above characterizations of top-, mid-, and bottom-level ontologies. Members of the top-level architecture are top-level ontologies designed to be domain-neutral in the sense that the ontologies in question are "created to represent the categories that are shared across a maximally broad range of domains" [2]. Example top-level ontologies include Basic Formal Ontology [3, 15] and TUpper [16]. Members of the domain-level architecture are ontologies designed to represent entities within some specified domain, thereby using fine-grained terms and relational expressions. Examples are the Neurological Disease Ontology [17] and the Cyber Ontology [18]. Members of the mid-level architecture (or "middle architecture") are designed to represent entities at a level of generality lower than those in the top-level architecture and a higher than those in the domain-level. Example mid-level ontologies include the Industrial Ontologies Foundry Core (IOFC) [6] and the Common Core Ontologies (CCO) suite [19, 20]. Building on these architectures, we defend individually necessary and jointly sufficient criteria for membership in the middle architecture, arguing that members warrant being counted as mid-level ontologies.

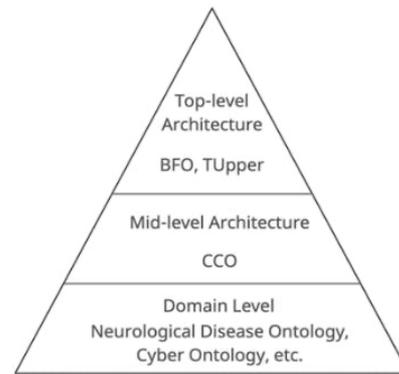

**Figure 1.** Top, middle, and domain architecture layers.

## 2. Considerations of Scope

Though our goal is to provide criteria for membership in the middle architecture, it is useful to begin by engaging with historical characterizations of mid-level ontologies. Most of these simply describe mid-level ontologies as sitting between top- and domain-level ontologies [1, 4, 5, 6]; but some have attempted to define mid-level ontologies directly [10, 11]. A theme in all of these contributions, whether implicit or explicit, is the notion of ontology *scope*, or what an ontology is meant to represent. For example, the scope of the Cyber Ontology is "entities relevant to the digitization, manipulation, and transfer of information using telecommunication networks, especially as they pertain to activities in cyberspace." [18]

The scope of a given ontology may be understood along both vertical and horizontal axes. *Vertical scope* is composed of, on the one hand, the most general and the least general groupings of entities in an ontology taxonomy – what we call the *upper* and *lower bound*, respectively. In this parlance, an upper bound for a top-level ontology such as BFO is represented by the class

---

[3] Our characterization is intended to be close to the "ontologi*cal* architecture" of [1] and "ontology architecture" of [4], [12], and [13]. But see [14] where the former is used to describe the structure of specific ontologies.

'entity' [3] while a lower bound is represented by the class 'object', which has no further refinements within BFO, but is a starting point for numerous BFO extension ontologies [21].

*Horizontal scope* reflects the intended breadth of entities covered by an ontology. To illustrate, a domain-level ontology implemented using terms and relational expressions that track as closely as possible entities in the relevant domain [22] will exhibit a horizontal scope delimited by the domain itself. One would not expect instances of airplanes or soccer matches to be within the purview of, say, the Cyber Ontology. A top-level ontology like BFO provides an example of a rather wide horizontal scope, namely *everything that exists*. This is indeed characteristic of top-level ontologies which satisfy the ISO/IEC 21838:1 Top-Level Ontologies Part 1: Requirements [2].

There is an intimate connection between upper bound and horizontal scopes, in that entities composing the horizontal scope of an ontology should be reflected in its most general groupings. The upper bound of BFO – reflected by the class 'entity' – aligns with its horizontal scope – everything that exists, has existed, or will exist – and indeed, the latter is reflected in the BFO definition of 'entity' [3]. Upper bounds and horizontal scope need not always align, as when an ontology is designed with a horizontal scope that is not sufficiently matched by its most general groupings. This may occur when, for example, an ontology is developed for a specific domain with a limited horizontal scope, but later expands that scope without reflecting this expansion in its upper bound. Alternatively, upper bounds and horizontal scope may come apart when placeholder classes are introduced in an ontology to signal its upper bound, but without the intention to represent entities falling under those classes. Neither scenario reflects ontology engineering best practices, suggesting that upper bounds and horizontal scopes should align.

When a domain ontology extends downwards from an ontology containing more general terms, the domain ontology should exhibit horizontal scope based on the relevant domain and to the extent possible exhibit an upper bound based on the lower bound of the higher-level ontology. If the domain ontology is sufficiently fine-grained, it should exhibit a clear lower bound as well. For example, the Occupation Ontology (OccO) [23] is a domain ontology developed to integrate data concerning occupation classification codes, such as the UK National Statistics Standard Occupational Classification (UK SOC) [24], and the European Skills, Competences, Qualifications and Occupations (ESCO) [25]. Because OccO adopts BFO and its design principles, OccO contains a clearly defined upper bound drawn from the lower bound of BFO, reflected in OccO's most general classes such as 'occupation role' extending from classes in BFO's lower bound such as 'role'. Because OccO is circumscribed to represent occupation classification codes, it exhibits a clearly defined horizontal scope. Because OccO is not intended to be developed below the level of generality needed to represent occupation codes, it contains clear lower bounds as well.

By way of another illustration, when domain ontologies are developed to directly reflect database structures representing a given domain, they may exhibit clear horizontal, upper, and lower bounds reflected by the boundaries of the database structure itself. For example, a relational database representing usernames and passwords that is transformed into a corresponding ontology may have bounds identifiable in the column headers and cells extracted from the database. Many ontologies developed following the so-called "bottom-up strategy" exhibit upper and lower bounds, and horizontal scope, insofar as they are primarily designed to represent exactly one clearly circumscribed domain [26].

## 3. Middle Architecture

We define 'middle architecture' in such a way that it consists solely of mid-level ontologies. Vertical and horizontal scope provide lines along which to identify necessary and sufficient criteria characterizing an architecture of this type.

## 3.1. The Extend Constraint

Ontologies in the middle architecture are 'middle' with respect to some ontology in the top-level architecture. We leverage criteria for inclusion in the top-level architecture from ISO/IEC 21838:1 Top-Level Ontologies—Requirements [2]. These ontologies are designed to represent categories, or general classes across a maximally broad range of domains. We maintain that ontologies satisfying the requirements of 21838:1 count as members of the top-level architecture. Moreover, ontologies in the middle architecture must extend from a top-level ontology thus defined.[4] We codify this as follows:

> **EXTEND** Middle architecture ontologies extend from at least one ontology satisfying the requirements specified in ISO/IEC 21838:1.

Where an ontology $O$ extends ontology $O^*$ when $O^*$ is a refinement of the intended interpretation of $O$ that is achieved by adding new class vocabulary to $O$. **EXTEND** enforces an upper bound for middle architecture ontologies. For example, the CCO suite consists of 11 ontologies[5] (**Figure 2**) and each extends from one or more classes in BFO. The most general classes in each of these extensions of BFO collectively represent the upper bound for CCO [19, 20], examples being 'agent' and 'artifact'.

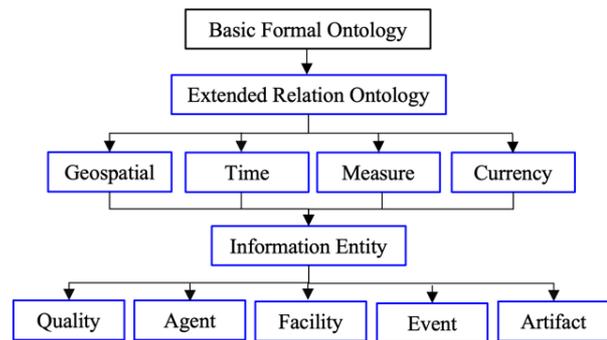

**Figure 2.** Basic Formal Ontology and the Common Core Ontologies Suite

Two points are worth emphasizing: First, by **EXTEND** a mid-level ontology that does not extend from a top-level ontology satisfying 21838:1 is not in the middle architecture as we define it.[6] Second, **EXTEND** does not exclude middle architecture ontologies that extend from multiple top-level ontologies, as long as at least one of the parent ontologies satisfies the requirements in 21838:1.[7]

## 3.2. The Delimit Constraint

We maintain that middle architecture ontologies should themselves be composed of content that is defined using terms and relational expressions extending ultimately from the vocabulary of the relevant top-level ontology referenced in **EXTEND**. This should be no surprise, as implemented ontologies that import a top-level ontology often do so in the interest of creating child classes or relations in just this manner.

Moreover, we maintain that middle architecture ontologies must be composed *only* of content based on the top-level referenced in **EXTEND**. This is less contentious than it sounds if we remember to keep separate *ontologies as intended semantics*[8] from *implementations of ontologies*. Ontologies in the former sense may be implemented in one or more formal languages, where an implementation is meant to reflect the intended interpretation of that ontology using just one specific formal language.[9] Formal language options for implementations include the Common Logic Interchange Format (CLIF) [28] and the Web Ontology Language (OWL2) [29]. While some researchers seem to suggest ontologies are equivalent to their implementations, i.e. by suggesting ontologies are formal theories [1, 30, 31], such claims lead rather quickly to puzzles. An OWL2 implementation of an ontology intended to represent the Allen Interval Algebra [33] will be unable to do so owing to OWL2's constraint on non-simple properties; in contrast, an implementation of the ontology in the more expressive CLIF might capture such an intended interpretation. Importantly, each would be an implementation of the same ontology. Ontologies are closer to intended semantics than to files stored in repositories.

---

[4] Note, requiring middle architecture ontologies extend from some ISO/IEC 21838:1 top-level ontology does not require that any specific top-level ontology must be used.
[5] CCO extensions exist, such as the Modal Relations Ontology (MRO) [19]. These are not, however, intended to be or to be part of some mid-level ontology suite.
[6] Similarly, we acknowledge there are top-level ontologies that do not inhabit the top architecture; we are only committed to any inhabitant of this architecture being a top-level ontology.
[7] We are not, however, asserting that the ontology in question cannot be a mid-level ontology. We return to this point at the close of **3.4**.
[8] As in [3], ontologies are "representational artifact whose representations are intended to designate some combination of universals, defined classes, and certain relations between them."

Our assumption, then, is that middle architecture ontologies exhibit intended semantics that are based on and only based on the intended semantics of a top-level from which they extend. Of course, implementations of ontologies leveraging BFO as a top-level sometimes include, for example, classes that suggest there are siblings of the most general BFO class 'entity'. This should not, however, by itself rule out a putative mid-level ontology with this feature from membership in the middle architecture. That determination is made with respect to the intended semantics of the mid-level ontology. This discussion justifies the following constraint, namely:

> **DELIMIT** Middle architecture ontologies are composed of all and only content ultimately extended from the upper bound of the top-level ontology referenced in **EXTEND**.

To illustrate what we mean by 'ultimately extended', consider the OWL2 implementation of CCO, which contains the class 'measurement unit'. This class is not an *immediate* owl:subClassOf of 'entity' in BFO but is connected to 'entity' through a series of owl:subClassOf relations. In that sense, 'measurement unit' ultimately extends from 'entity'.

**EXTEND** and **DELIMIT** enforce a specific type of upper bounds for middle architecture ontologies; a natural next step would be to identify a criterion for middle architecture ontology lower bounds. **EXTEND** and **DELIMIT** undermine one possible strategy which attempts to specify a criterion for determining mid-level lower bounds in general [10] that can be applied to determine middle architecture ontology lower bounds in particular:

> **(*)** For a given ontology element *e*, natural number $n > 1$, and distinct domain-level ontologies $o_1...o_n$: If *e* is appropriately reused in $o_1...o_n$ then the primary residence of *e* should be a more general ontology imported by $o_1...o_n$.

**(*)** is, in certain circumstances, a useful principle. Consider that the term 'infection' is plausibly used across all infectious disease ontologies. Housing the term 'infection' term in, say, an ontology whose scope is restricted to influenza would require other infectious disease ontologies to import 'infection' from that influenza ontology. Better to place 'infection' in a more general ontology, such as the Infectious Disease Ontology (IDO) [29], alongside terms commonly used across multiple infectious disease domain ontologies. **(*)** justifies such a decision. The thought is that if **(*)** can provide a dividing line between mid- and domain-level ontologies generally, then it can provide a lower bound for middle architecture ontologies, in particular.

Unfortunately, because domain ontologies that extend from the same top- or mid-level ontology may legitimately represent the same domain in different ways, **(*)** fails to provide a defensible cutoff between mid- and domain-level ontologies; hence, **(*)** cannot be leveraged in our criteria. Consider that a domain ontology intended to represent car accidents represents a domain that plausibly overlaps with the car insurance domain just as well as do domain ontologies built specifically to represent car insurance. Both ontologies may plausibly include a class 'Honda Civic' but this should not entail that 'Honda Civic' is a class that belongs in a mid-level ontology. Similarly, a domain ontology representing strategies for recycling vehicles might also have need for 'Honda Civic' within its scope. But three domain ontologies using 'Honda Civic' should not force this class into the mid-level. One might still be tempted to claim that for some sufficiently large *n*, reuse across *n* domain ontologies warrants inclusion in a mid-level ontology. However, because mid-level ontologies can be extended by overlapping but distinct domain-level ontologies in potentially infinite ways, leveraging **(\*)** – even for some large *n* – to provide a firm cutoff between the mid- and domain ontologies runs the risk of collapsing the corresponding architectures.

---

[9] Compare [27] in which ontologies are described as documents that are "realized in" document versions.

It is unclear how to identify a defensible lower bound for mid-level ontologies; it is similarly unclear how to identify such a lower bound for middle architecture ontologies. While rules of thumb have been suggested – such as limiting the number of subclasses of a given mid-level ontology to no more than three [10] – such rules are arbitrary. Rather than attempting to identify a firm cutoff, we propose that we rely instead on existing consensus regarding mid-level ontology content. There is often much more agreement as to what should be included in a given mid-level ontology than there is disagreement. For every contentious, potentially borderline class or relation in CCO implementations – such as 'flywheel' or 'is_first_cousin_of' – there are many more uncontentious classes – such as 'agent', 'artifact', 'information content entity', 'measurement', and 'is_about', to name a few [20]. Most importantly, we should not take the lack of a firm cutoff for what should and should not be included in a mid-level ontology to undermine the project of identifying criteria for middle architecture membership.

### 3.3. The Hub Constraint

While there are examples of mid-level ontologies [5] intended to be implemented solely as single artifacts, we should not expect all middle architecture ontologies to be similarly structured. That is, we should permit, under certain conditions, collections of ontologies in the middle architecture, even when no single member would count as a mid-level ontology.

To codify this point, we expand upon the notion of *ontology modules*.[10] Ontology modules are standardly characterized as self-contained components of larger ontologies, able to be integrated with other self-contained components of larger ontologies [33]. Building on this characterization we introduce *ontology hubs* as: ontology modules designed to serve as foundations from which more specific ontologies – *ontology spokes* – extend [34]. As an example, creation of IDO [35] spurred development of extension ontologies covering brucellosis [36], influenza [37], and coronavirus [38], among others. Ontologies representing standard groupings of pathogens, e.g. parasite, bacteria, fungus, virus, share significant content in common, warranting the creation of ontology modules, such as the Virus Infectious Disease Ontology [39], an ontology hub for virus ontology spokes, like the Coronavirus Infectious Disease Ontology [38].

Ontology hubs provide the lines along which to make sense of a collection of ontologies being a member of the middle architecture. At a minimum, middle architecture ontologies – whether unified ontologies or collections – should be composed of ontology hubs, none of which are members of the top-level architecture. We go further, however, in maintaining that middle architecture ontologies should *only* be composed of such ontology hubs. Such a constraint is intended to exclude from the middle architecture collections of ontology modules combined with either top- or domain-level ontologies. For example, the result of combining CCO and OccO would be an ontology outside the middle architecture since the latter is not designed to serve as a foundation for ontology spokes; similarly, the result of combining CCO with an ontology hub of BFO that satisfies the criteria of 21838:1 would not count as a member of the middle architecture since the latter hub would be a member of the top-level architecture. The idea of ontology spokes as described thus reflects the intuition that mid-level ontologies are more general than domain-level ontologies but less general than top-level ontologies. We need only qualify our ontology hub requirement to explicitly rule out overlapping scope to provide our next criterion:

> **HUB** Middle architecture ontologies are composed of all and only ontology hubs none of which overlap in scope with any other.[11]

---

[10] Focusing on implementations: if $O=(C, R)$ represents an ontology vocabulary, $C=\{c_1, c_2, ...,c_n\}$ terms, and $R=\{R_1(c_i, c_j) ... R_m(c_x, c_y)\}$ relations, then an ontology module $O_M = (C_M, R_M)$ of $O$ is such that $C_M \subset C$ and $R_M \subset R$. Compare [33].

As a limit case, **HUB** can be satisfied by a single ontology hub. More generally, **HUB** may be satisfied by a collection of one or more ontology hubs. For example, as indicated earlier, CCO is composed of 11 ontology hubs together designed to exhaust the scope of BFO, with each hub covering some broad domain of interest, such as information or artifacts.[12]

### 3.4. The Inheritance Constraint

Middle architecture ontologies should exhibit a tight connection with the top-level ontologies which they extend by inheriting their horizontal scope. For example, if a given middle architecture ontology extends BFO, then it should have as its horizontal scope what BFO is designed to cover, namely, everything. It is worth noting that such a commitment conflicts with characterizations of "mid-level ontologies" as ontologies "that represent relatively general categories common to many domains of interest." [11] One way to interpret this characterization is to understand "mid-level ontology" as picking out ontologies representing some broad user community or perhaps scientific field, such as biomedicine, manufacturing, education, and so forth. On such a picture, a given top-level ontology might be extended by both a biomedical mid-level ontology, a distinct manufacturing mid-level ontology, a distinct education mid-level ontology, and so on. Call these *relative mid-level ontologies*.

Relative mid-level ontologies are not suitable members of the middle architecture. We are committed to minimizing *scope creep* [3] among middle architecture ontologies; the most plausible way to do so is to require that middle architecture ontologies inherit the horizontal scope of their top-level. Scope creep emerges when an ontology intended to represent some specific domain is constructed with insufficient foresight, so that it later needs to be expanded beyond that domain. Consider the Industrial Ontologies Foundry Core (IOFC), described by its developers as a mid-level ontology with respect to industrial manufacturing and services [6]. IOFC extends directly from BFO and so inherits its minimal top-level terms and relational expressions. Accordingly, IOFC developers found a need to mint new ontology vocabulary representing agents, artifacts, information, and so on, much of which was outside the scope of IOFC proper.[13] Observe that a natural antidote to the preceding scope creep would be to store relevant terms and relational expressions representing artifacts, information, etc. needed by the IOFC relative mid-level ontology in a 'more general' mid-level ontology which IOFC imports. Scope creep is, however, pervasive among relative mid-level ontologies [3, 26].[14] With enough relative mid-level ontologies aiming to avoid scope creep there would be pressure to create a 'most general mid-level ontology'. Because scope creep is notoriously challenging to address once established, our criteria should encourage *starting* with such a 'most general mid-level ontology'. In other words, to avoid scope creep, we should encourage middle architecture ontologies to inherit the horizontal scope of the top-level ontology from which they extend.

---

[11] Compare [40] where it is argued that OBO Foundry ontologies should have orthogonal scope.
[12] Note that because we restrict our focus to ontology hubs outside the top-level architecture, middle architecture ontologies cannot be collections of top-level ontologies.
[13] Similarly for OBO Foundry [40] ontologies extending BFO that have minted ontology terms and relational expressions representing artifacts, information, etc., none of which are interoperable with those of IOFC.
[14] See several examples in **Section 4** below.

Perhaps more contentiously, we maintain that middle architecture ontologies should be designed to inherit that scope by introducing more specific ontology content. As a first pass:

> **(\*\*)** Middle architecture ontologies must contain at least one subclass for each class reflecting the lower bound of the top-level ontology they extend.

For example, BFO classes such as 'function' and 'history' are extended in CCO to 'artifact function' and 'artifact history', respectively. While **(\*\*)** seems initially attractive, it is revealed on reflection to be too strong. Consider that subclasses of BFO's 'spatial region' are still rarely, if ever, introduced correctly [41]. For example, CCO currently includes subclasses for 'one-dimensional spatial region' [19] such as 'Coordinate System Axis', which is a "A One-Dimensional Spatial Region defined by a Coordinate System for the purpose of identifying the position of entities along one dimension of the Coordinate System's spatial framework." Here we see conditions for counting as a 'one-dimensional spatial region' given entirely in terms of information *about* [43] spatial regions, namely, coordinate systems which are themselves subclasses of 'information content entities' in CCO. This is common among subclasses of BFO's 'spatial region'. We maintain such subclasses should be deprecated. Extensions of child classes of 'spatial region' will, we believe, not be needed by most ontologies. More generally, it seems plausible that some 21838:1 top-level ontology will include classes that should not be extended by middle architecture ontologies. Hence, **(\*\*)** is too strong.

There is a more flexible path forward that leverages requirements outlined in 21838:1. Any top-level ontology satisfying this standard must provide explanations for how data across the breadth areas in **Table 1** will be represented.

Table 1. Breadth Areas for 21838:1 Top-Level Ontologies.

| | |
|---|---|
| Space and Time | Qualities and other Attributes |
| Actuality and Possibility | Quantities and Mathematical Entities |
| Classes and Types | Processes and Events |
| Time and Change | Constitution |
| Parts, Wholes, Unity, Boundaries | Causality |
| Space and Place | Information and Reference |
| Scale and Granularity | Artifacts, Socially Constructed Entities |
| Mental entities, imagined entities, fiction, mythology, and religion | |

These breadth areas provide guidance to those who intend to develop or evaluate top-level ontologies with respect to the range of types of data they can represent. Strictly speaking, top-level ontologies satisfying 21838:1 need not in every case even "include classes or types that cover one or more of the areas identified". In cases where a putative top-level ontology does not do so, it must document how it will address such coverage, perhaps by referencing other, external ontologies that extend the top-level. For example, BFO satisfies coverage of information artifacts in ISO/IEC 21838:2 [15], by referencing CCO's Information Entity Ontology and the treatment of information artifacts therein in terms of the BFO class 'generically dependent continuant'. We leverage these breadth areas to provide a constraint on middle architecture ontologies that is more flexible than **(\*\*)**:[15]

---

[15] Note, satisfying **(\*\*)** is one way to satisfy **INHERITANCE**, though not the only way.

**INHERITANCE** Middle architecture ontologies are composed of all and only content extended from each breadth area of the top-level ontology referenced in **EXTEND**.

By requiring that middle architecture ontologies extend from each breadth area in **Table 1**, rather than each grouping of a top-level ontology lower bound, we avoid forcing the creation of unhelpful and potentially confused classes just to satisfy our constraints. We maintain a firmer position than 21838:1 insofar as middle architecture ontologies cannot satisfy **INHERITANCE** by simply documenting how they can be extended to cover each breadth area, even if that documentation references external extension ontologies. Rather, to satisfy **INHERITANCE** middle architecture ontologies must explicitly represent each breadth area.

Observe **EXTEND**, **HUB**, and **INHERITANCE** entail that a middle architecture ontology consisting of *two or more* ontology hubs cannot extend distinct 21838:1 top-level ontologies. By **INHERITANCE**, the ontology hubs must contain at least one subclass for each breadth area of each top-level ontology referenced by **EXTEND**. Because the top-level ontologies exhibit overlapping scope, so will the ontology hubs, violating **HUB**.[16]

## 4. Applying the Criteria

**Table 2** summarizes our criteria for membership in the middle architecture. We turn now to evaluating those ontologies which are potential members of the middle architecture.

**Table 2.** Individually Necessary and Jointly Sufficient Criteria for the Middle Architecture.

| | |
|---|---|
| **EXTEND** | Middle architecture ontologies extend from at least one ontology satisfying the requirements specified in ISO/IEC 21838:1. |
| **DELIMIT** | Middle architecture ontologies are composed of all and only content ultimately extended from the upper bound of the top-level ontology referenced in **EXTEND**. |
| **HUB** | Middle architecture ontologies are composed of all and only ontology hubs none of which overlap in scope with any other. |
| **INHERITANCE** | Middle architecture ontologies are composed of all and only content extended from each breadth area of the top-level ontology referenced in **EXTEND**. |

We have used CCO as our running example, so it should be no surprise that it satisfies the criteria. The 11 ontologies comprising the CCO suite are disjoint ontology hubs, thus satisfying **HUB**. CCO adopts BFO as a top-level ontology, thus satisfying **EXTEND**. CCO extends ultimately from BFO's breadth areas, satisfying **INHERITANCE**; but CCO does not include among the 11 modules any class that extends outside the scope of BFO, thus satisfying **DELIMIT**. By these criteria, the 11 ontologies that compose the CCO suite count as a middle architecture ontology.

The Ontology for Biomedical Investigations (OBI) [44] grew out of various OBO Foundry efforts. Accordingly, OBI reused ontologies developed for many domains of interest across the OBO community. OBI adopts BFO as a top-level, thus satisfying **EXTEND**; it is arguably a single ontology hub, thus satisfying **HUB**; and it does not include any class that extend beyond the scope of BFO, thus satisfying **DELIMIT**. OBI does not, however, cover all breadth areas identified in 21838:1 and leveraged in **INHERITANCE**. For example, the scope of OBI is not intended to cover imagined entities, fiction, mythology, and religion. Hence, according to our criteria OBI is not a middle architecture ontology. This is to take nothing away from OBI, however. OBI is simply not a mid-level ontology of the sort we are interested in here.

---

[16] The present criteria do not rule out a *single* middle architecture ontology extending from two or more top-level ontologies satisfying 21838:1.

The Industrial Ontologies Foundry Core (IOFC) [6] was developed to provide terminological integration for BFO-compliant ontologies covering the domains of industrial manufacturing, service, and maintenance. Because IOFC adopts BFO as a top-level ontology, it satisfies **EXTEND**. Moreover, IOFC is a single ontology hub, thus satisfying **HUB**, and does not extend outside the scope of BFO, thus satisfying **DELIMIT**. As with OBI, however, IOFC does not satisfy **INHERITANCE** given the limitations of its scope to industrial manufacturing, e.g. IOFC is not designed to cover boundaries, space and time, or fiction breadth areas. Hence, IOFC is not a middle architecture ontology.

The authors of the present article are members of the Buffalo Toronto Ontology Alliance (BoaT) and have worked with members of the Toronto Virtual Enterprise (TOVE) community to align "their respective suites of ontologies." [44] The Toronto Virtual Enterprise (TOVE) project [45] aims to promote data-driven city policy making by leveraging ontologies. To our knowledge, no single ontology or combination of ontologies in the TOVE suite is intended to count as a mid-level. Nevertheless, given the breadth covered by TOVE ontologies – spanning a range of domains such as activities, resources, and time – it is instructive to explore the extent to which a *collection* of TOVE ontologies may count as a middle architecture ontology. The most general TOVE ontologies are properly modularized ontologies which avoid overlapping scope, and so satisfy **HUB**. Nevertheless, the ontologies do not as yet adopt any top-level ontology, and thus do not satisfy **EXTEND**, **DELIMIT**, or **INHERITANCE**. That said, given the breadth of coverage and careful engineering, some combination of the highest-level TOVE ontologies could plausibly count as a middle architecture ontology, once properly arranged under a top-level ontology satisfying 21838:1.

## 5. Conclusion

Ontologies can be characterized along levels of generality. The purpose of a well-developed mid-level ontology is to provide a foundation of ontology elements more specific than a top-level ontology but more general than any domain ontology. A mid-level ontology should offer a connection between top- and domain-level ontologies, and so – we maintain – facilitate the development of ontologies following the so-called "middle-in strategy" [26]. Given the recent interest in mid-level ontologies by established groups such as the IEEE [9], providing criteria for their identification will set standards for future ontology development. We have thus introduced four criteria - **EXTEND**, **DELIMIT**, **HUB**, and **INHERITANCE** - characterizing the middle architecture, which consists solely of mid-level ontologies, while arguing against criteria such as **(\*)** and **(\*\*)**. On our proposal, membership in the middle architecture requires consisting of one or more non-overlapping ontology hubs, which extend all breadth areas of a 21838:1 top-level ontology.

## Acknowledgements

Thanks to participants of the 2023 University at Buffalo Fall Ontology Sprint for feedback on early versions of the criteria: Alec Sculley, Ji Soo Seo, Federico Donato, Sean Kindya, Giorgio Ubbiali, James Egan, Adam Taylor, Hector Guzman-Orozco, and Michael Rabenberg. Many thanks to the IEEE P3195 Mid-Level Ontology and Extensions Working Group for discussion of these criteria: Alan Ruttenberg, Brian Haugh, Alex Cox, Neil Otte, Cameron More, Forrest Hare, Alan Belasco, Austin Leibers, Eric Merrell, Tim Prudhomme, Jonathan Vajda, Steven Wartik, and Jim Schoening. Special thanks to Adrien Barton and Fumiaki Toyoshima for extensive feedback on an early version of this paper.